\documentclass[a4paper]{article}

\usepackage[utf8]{inputenc}
\usepackage{INTERSPEECH_v2}
\usepackage{float}
\usepackage{hyperref}
\usepackage{enumitem}

\hypersetup{
    colorlinks,
    linkcolor={blue!20!black},
    citecolor={blue!20!black},
    urlcolor={blue!20!black}
}

\usepackage{amsmath,amssymb}
\DeclareMathOperator{\E}{\mathbb{E}}
\usepackage{xcolor}

\usepackage{mathrsfs}
\newcommand{\ve}[1]{\textbf{#1}}		% vectors
\newcommand{\di}[1]{\mathcal{#1}}		% distributions
\usepackage{url}
\usepackage{todonotes}

\usepackage{pgfplots}
\usepgfplotslibrary{statistics}

\title{SEGAN: Speech Enhancement Generative Adversarial Network}
\name{Santiago Pascual$^1$, Antonio Bonafonte$^1$, Joan Serr{\`a}$^2$}
\address{
  $^1$Universitat Polit{\`e}cnica de Catalunya, Barcelona, Spain\\
  $^2$Telef{\'o}nica Research, Barcelona, Spain}
\email{santi.pascual@upc.edu, antonio.bonafonte@upc.edu joan.serra@telefonica.com}

\pgfplotsset{compat=1.14}

\begin{document}

\maketitle

\begin{abstract}
%Speech enhancement is about increasing speech signals' intelligibility, as well as their perceptual quality. 
%This task can precede many others in the speech processing domain, where a clean signal is rather preferred to achieve better detections, as in automatic speech recognition, or a higher quality acoustic modeling, as in text-to-speech.
Current speech enhancement techniques operate on the spectral domain and/or exploit some higher-level feature. The majority of them tackle a limited number of noise conditions and rely on first-order statistics. To circumvent these issues, deep networks are being increasingly used, thanks to their ability to learn complex functions from large example sets. %Besides, deep networks are known to effectively deal with structured and correlated data like speech, without any need for hand-crafted feature transformations, so that models can be built within an end-to-end framework. 
In this work, we propose the use of generative adversarial networks for speech enhancement. In contrast to current techniques, we operate at the waveform level, training the model end-to-end, and incorporate 28~speakers and 40~different noise conditions into the same model, such that model parameters are shared across them. %The proposed model can be seen as a learned loss function within an adversarial framework that works at the waveform level. 
We evaluate the proposed model using an independent, unseen test set with two speakers and 20~alternative noise conditions. The enhanced samples confirm the viability of the proposed model, and both objective and subjective evaluations confirm the effectiveness of it. 
%We evaluate the proposed approach using an independent test set of xx~hours of audio and yy~noise conditions and perform both objective and subjective evaluations, showing a xx~PesQ improvement with respect to xxx and a strong preference for the proposed model in the subjective tests.
With that, we open the exploration of generative architectures for speech enhancement, which may progressively incorporate further speech-centric design choices to improve their performance.
\end{abstract}


\section{Introduction}

Speech enhancement tries to improve the intelligibility and quality of speech contaminated by additive noise~\cite{Loizou2013Book}. Its main applications are related to improving the quality of mobile communications in noisy environments. However, we also find important applications related to hearing aids and cochlear implants, where enhancing the signal before amplification can significantly reduce discomfort and increase intelligibility~\cite{yang2005}. Speech enhancement has also been successfully applied as a preprocessing stage in speech recognition and speaker identification systems~\cite{yu2008,maas2012,ortega96icslp}. 

Classic speech enhancement methods are spectral subtraction~\cite{berouti79}, Wiener filtering~\cite{Lim78}, statistical model-based methods~\cite{ephraim92}, and subspace algorithms~\cite{dendrinos91,ephraim95}. Neural networks have been also applied to speech enhancement since the 80s~\cite{Tamura88ICASSP,Parveen04ICASSP}. %Recently, the architectures more widely adopted for denoising are refereed to as Denoising Auto-Encoders (DAE)~\cite{lu2013}. A DAE is a Neural Network which attempts to map noisy inputs to their clean version. 
Recently, the denoising auto-encoder architecture~\cite{lu2013} has been widely adopted. However, recurrent neural networks (RNNs) are also used. For instance, the recurrent denoising auto-encoder has 
%Recurrent Neural Networks are also used for denoising task. The architecture based on Recurrent Neural Networks are called Recurrent Denoising Auto-Encoders (RDAE) and they have 
shown significant performance exploiting the temporal context information in embedded signals. Most recent approaches apply long short-term memory networks to the denoising task~\cite{maas2012,Weninger15LVASS}. In~\cite{Xu15TASLP} and~\cite{Kumar16IS}, noise features are estimated and included in the input features of deep neural networks. The use of dropout, post-filtering, and perceptually motivated metrics are shown to be effective. 

Most of the current systems are based on the short-time Fourier analysis/synthesis framework~\cite{Loizou2013Book}. They only modify the spectrum magnitude, as it is often claimed that short-time phase is not important for speech enhancement~\cite{wang82}. However, further studies~\cite{paliwal2011} show that significant improvements of speech quality are possible, especially when a clean phase spectrum is known. In 1988, Tamura et~al.~\cite{Tamura88ICASSP} proposed a deep network that worked directly on the raw audio waveform, but they used feed-forward layers that worked frame-by-frame (60~samples) on a speaker-dependent and isolated-word database.

%The proposed architecture is based on the adversarial training technique, where the Generator (G) network learns to clean a full chunk of waveform in a single inference pass, whilst the Discriminator (D) network tries to identify whether the waveform comes from G or from the training set.
A recent breakthrough in the deep learning generative modeling field are generative adversarial networks (GANs)~\cite{goodfellow2014generative}. GANs have achieved a good level of success in the computer vision field to generate realistic images and generalize well to pixel-wise, complex (high-dimensional) distributions~\cite{Isola16ARXIV,Mao16ARXIV,radford2015unsupervised}. As far as we are concerned, GANs have not yet been applied to any speech generation nor enhancement task, so this is the first approach to use the adversarial framework to generate speech signals. 

The main advantages of the proposed speech enhancement GAN (SEGAN) are:
\begin{itemize}[noitemsep,topsep=0pt,parsep=0pt,partopsep=0pt,leftmargin=15pt]
\item It provides a quick enhancement process. No causality is required and, hence, there is no recursive operation like in RNNs.
\item It works end-to-end, with the raw audio. Therefore, no hand-crafted features are extracted and, with that, no explicit assumptions about the raw data are done.
\item It learns from different speakers and noise types, and incorporates them together into the same shared parametrization. This makes the system simple and generalizable in those dimensions.
\end{itemize}

%The paper is structured as follows. In Sec.~\ref{sec:gan}, we give an overview of GANs. This is followed by the description of our proposed model in Sec.~\ref{sec:segan} and its experimental setup in Sec.~\ref{sec:experimental setup}. The results are shown in Sec.~\ref{sec:results}, and conclusions are discussed in Sec.~\ref{sec:conclusions}.
In the following, we give an overview of GANs (Sec.~\ref{sec:gan}). Next, we describe the proposed model (Sec.~\ref{sec:segan}) and its experimental setup (Sec.~\ref{sec:experimental setup}). We finally report the results (Sec.~\ref{sec:results}) and discuss some conclusions (Sec.~\ref{sec:conclusions}).

\section{Generative Adversarial Networks}
\label{sec:gan}

GANs~\cite{goodfellow2014generative} are generative models that learn to map samples $\ve{z}$ from some prior distribution $\di{Z}$ to samples $\ve{x}$ from another distribution $\di{X}$, which is the one of the training examples (e.g.,~images, audio, etc.). The component within the GAN structure that performs the mapping is called the generator (G), and its main task is to learn an effective mapping that can imitate the real data distribution to generate novel samples related to those of the training set. Importantly, G does so not by memorizing input-output pairs, but by mapping the data distribution characteristics to the manifold defined in our prior $\di{Z}$. 

\begin{figure}[t!]
\centering
\includegraphics[width=1\linewidth]{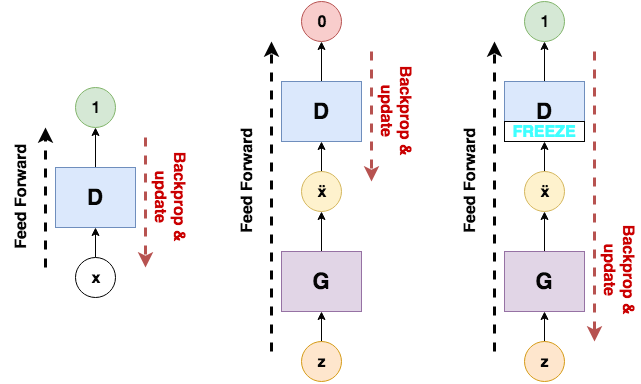}
\caption{\label{fig:GAN}GAN training process. First, D back-props a batch of real examples. Then, D back-props a batch of fake examples that come from G, and classifies them as fake. Finally, D's parameters are frozen and G back-props to make D misclassify.}
\end{figure}

The way in which G learns to do the mapping is by means of an adversarial training, where we have another component, called the discriminator (D). D is typically a binary classifier, and its inputs are either real samples, coming from the dataset that G is imitating, or fake samples, made up by G. The adversarial characteristic comes from the fact that D has to classify the samples coming from $\di{X}$ as real, whereas the samples coming from G, $\di{\hat{X}}$, have to be classified as fake. This leads to G trying to fool D, and the way to do so is that G adapts its parameters such that D classifies G's output as real. %, and hence the fact that they are "fighting" as they have opposite objectives. 
During back-propagation, D gets better at finding realistic features in its input %, such that these can be leaked to G as the gradients flow through D, which in turn will correct its 
and, in turn, G corrects its parameters to move towards the real data manifold described by the training data (Fig.~\ref{fig:GAN}).  
This adversarial learning process is formulated as a minimax game between G and D, with the objective 
\begin{equation}
\begin{aligned}
  %\underset{G}\min~\underset{D}\max~V(D,G) & = \E_{x\sim p_{\text{data}}(x)}\left[\log D(\ve{x})\right] + \\
  %		& + \E_{z\sim p_{\ve{z}}(\ve{z})}\left[\log \left(1 - D\left(G\left(\ve{z}\right)\right)\right)\right] \\
  \underset{G}\min~\underset{D}\max~V(D,G) & = \E_{\ve{x}\sim p_{\text{data}}(\ve{x})}\left[\log D(\ve{x})\right] + \\
  		& + \E_{\ve{z}\sim p_{\ve{z}}(\ve{z})}\left[\log \left(1 - D\left(G\left(\ve{z}\right)\right)\right)\right]. 
  \label{eq:gan_minimax}
\end{aligned}
\end{equation}
%A description of the training process is depicted in Figure~\ref{fig:GAN}.

We can also work with a conditioned version of GANs, where we have some extra information in G and D to perform mapping and classification (see~\cite{Isola16ARXIV} and references therein). In that case, we may add some extra input $\ve{x}_c$, with which we change the objective function to
\begin{equation}
%  \underset{G}{\text{min}}\underset{D}{\text{max}}V(D,G) = \E_{x ~ p_{data}(x, x_c)}[\log D(\boldsymbol{x}, \boldsymbol{x_c})] + \E_{x~p_{data}(x_c), z~p_{z}(z)}[\log (1 - D(G(\boldsymbol{z}, \boldsymbol{x_c})))]
\begin{aligned}
  \underset{G}\min~&\underset{D}\max~V(D,G) = \E_{\ve{x},\ve{x}_c\sim p_{\text{data}}(\ve{x},\ve{x}_c)}\left[\log D(\ve{x},\ve{x}_c)\right] + \\
  		& + \E_{\ve{z}\sim p_{\ve{z}}(\ve{z}), \ve{x}_c\sim p_{\text{data}}(\ve{x}_c)}\left[\log \left(1 - D\left(G\left(\ve{z},\ve{x}_c\right), \ve{x}_c\right)\right)\right]. 
 \label{eq:cgan_minimax}
\end{aligned}
\end{equation}
%This is the underlying structure used in this work.

There have been recent improvements in the GAN methodology to stabilize training and increase the quality of the generated samples in G. For instance, the classic approach suffered from vanishing gradients due to the sigmoid cross-entropy loss used for training. To solve this, the least-squares GAN (LSGAN) approach~\cite{Mao16ARXIV} substitutes the cross-entropy loss by the least-squares function with binary coding (1 for real, 0 for fake). With this, the formulation in Eq.~\ref{eq:cgan_minimax} changes to
\begin{equation}
\begin{aligned}
  \underset{D}\min~V_{\text{LSGAN}}(D) & = \frac{1}{2}\E_{\ve{x},\ve{x}_c\sim p_{\text{data}}(\ve{x}, \ve{x}_c)}[(D(\ve{x},\ve{x}_c) - 1)^{2}] +\\
    & + \frac{1}{2}\E_{\ve{z}\sim p_{\ve{z}}(\ve{z}),\ve{x}_c\sim p_{\text{data}}(\ve{x}_c)}[D(G(\ve{z},\ve{x}_c),\ve{x}_c)^{2}] 
  \label{eq:lsgan_d}
\end{aligned}
\end{equation}
\begin{equation}
  \underset{G}\min~V_{\text{LSGAN}}(G) = \frac{1}{2}\E_{\ve{z}\sim p_{\ve{z}}(\ve{z}),\ve{x}_c\sim p_{\text{data}}(\ve{x}_c)}[(D(G(\ve{z},\ve{x}_c),\ve{x}_c) - 1)^{2}].
  \label{eq:lsgan_g}
\end{equation}

\section{Speech Enhancement GAN}
\label{sec:segan}

The enhancement problem is defined so that we have an input noisy signal $\tilde{\ve{x}}$ and we want to clean it to obtain the enhanced signal $\hat{\ve{x}}$. %During training, we have, for every noisy signal, its clean reference $\ve{x}$. %The model proposed in this work follows the conditioned Generative Adversarial approach described Sec.~\ref{sec:gan}. 
We propose to do so with a speech enhancement GAN (SEGAN). 
In our case, the G network performs the enhancement. Its inputs are the noisy speech signal $\tilde{\ve{x}}$ together with the latent representation $\ve{z}$, and its output is the enhanced version $\hat{\ve{x}} = G(\tilde{\ve{x}})$. We design G to be fully convolutional, so that there are no dense layers at all. This enforces the network to focus on temporally-close correlations in the input signal and throughout the whole layering process. %, exploiting local relationships that could otherwise be lost with fully connected topologies. 
Furthermore, it reduces the number of training parameters and hence training time.

\begin{figure}[t]
\centering
\includegraphics[width=0.75\linewidth]{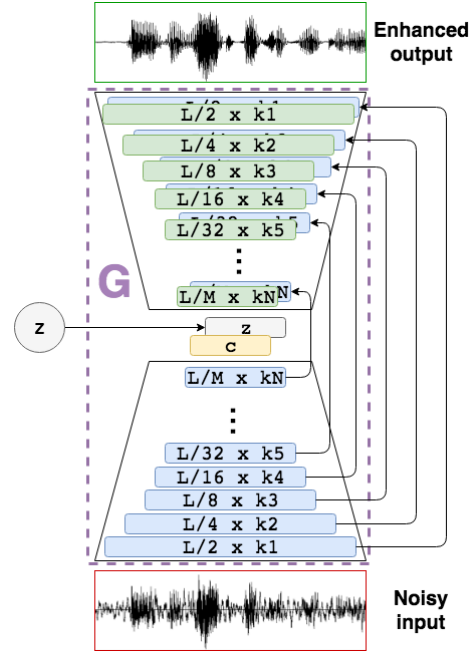}
\caption{\label{fig:GAE}Encoder-decoder architecture for speech enhancement (G network). 
%Feature maps $\boldsymbol{c}$ and $\boldsymbol{z}$ depict the thought vector and the prior sampling, respectively. %Both $\boldsymbol{c}$ and $\boldsymbol{z}$ have $\frac{L}{M}\times K\times N$ dimensions, concatenated in the channels dimension, so the final encoded representation is $\frac{L}{M}\times 2\times K\times N$. 
The arrows between encoder and decoder blocks denote skip connections. %$M$ and $N$ depend on the depth of our model.
}
\end{figure}

The G network is structured similarly to an auto-encoder (Fig.~\ref{fig:GAE}).  
%There are then two sections of the network: encoder and decoder. 
In the encoding stage, the input signal is projected and compressed through a number of strided convolutional layers followed by parametric rectified linear units (PReLUs)~\cite{he2015delving}, getting a convolution result out of every $N$ steps of the filter. We choose strided convolutions as they were shown to be more stable for GAN training than other pooling approaches~\cite{radford2015unsupervised}. Decimation is done until we get a condensed representation, called the thought vector $\ve{c}$, which gets concatenated with the latent vector $\ve{z}$. The encoding process is reversed in the decoding stage by means of fractional-strided transposed convolutions (sometimes called deconvolutions), followed again by PReLUs.%, which convert the hidden representation into the clean version of the input corrupted signal. 

The G network also features skip connections, connecting each encoding layer to its homologous decoding layer, and bypassing the compression performed in the middle of the model (Fig.~\ref{fig:GAE}). This is done because the input and output of the model share the same underlying structure, which is that of natural speech. Therefore, many low level details could be lost to reconstruct the speech waveform properly if we force all information to flow through the compression bottleneck. Skip connections directly pass the fine-grained information of the waveform to the decoding stage (e.g., phase, alignment). In addition, they offer a better training behavior, as the gradients can flow deeper through the whole structure~\cite{he2016deep}.

An important feature of G is its end-to-end structure, so that it processes raw speech sampled at 16\,kHz, getting rid of any intermediate transformations to extract acoustic features (contrasting to many common pipelines). In this type of model, we have to be careful with typical regression losses like mean absolute error or mean squared error, as noted in the raw speech generative model WaveNet~\cite{van2016wavenet}. These losses work under strong assumptions on how our output distribution is shaped and, therefore, impose important modeling limitations (like not allowing multi-modal distributions and biasing the predictions towards an average of all the possible predictions). Our solution to overcome these limitations is to use the generative adversarial setting. This way, D is in charge of transmitting information to G of what is real and what is fake, such that G can slightly correct its output waveform towards the realistic distribution, getting rid of the noisy signals as those are signaled to be fake. In this sense, D can be understood as learning some sort of loss for G's output to look real.%, so that the enhancement must respect the speech signal and eliminate all the surrounding noise as much as possible. %To achieve that, the training examples have to be informative of what we want to achieve, and therefore 

In preliminary experiments, we found it convenient to add a secondary component to the loss of G in order to minimize the distance between its generations and the clean examples. To measure such distance, we chose the $L_1$ norm, as it has been proven to be effective in the image manipulation domain~\cite{Isola16ARXIV, pathak2016context}. This way, we let the adversarial component to add more fine-grained and realistic results. The magnitude of the $L_1$ norm is controlled by a new hyper-parameter $\lambda$. Therefore, the G loss, which we choose to be the one of LSGAN (Eq.~\ref{eq:lsgan_g}), becomes
\begin{equation}
%  \underset{G}{\text{min}}V_{LSGAN}(G) = \frac{1}{2}(\E_{x~p_{data}(x_c), z~p_{z}(z)}[(D(G(\boldsymbol{z}, \hat{\boldsymbol{x}})) - 1)^{2}] + \lambda\cdot|(G(\boldsymbol{z}, \hat{\boldsymbol{x}}) - \boldsymbol{x})|
\begin{aligned}
  \underset{G}\min~V_{\text{LSGAN}}(G) & = \frac{1}{2}\E_{\ve{z}\sim p_{\ve{z}}(\ve{z}),\tilde{\ve{x}}\sim p_{\text{data}}(\tilde{\ve{x}})}[(D(G(\ve{z},\tilde{\ve{x}}), \tilde{\ve{x}}) - 1)^{2}] + \\
    & + \lambda~ \|G(\ve{z},\tilde{\ve{x}}) - \ve{x} \|_1.
  \label{eq:G_newloss}
\end{aligned}
\end{equation}

\section{Experimental Setup}
\label{sec:experimental setup}

%In this section the Database used for the experiments is described, as well as the concrete topology of the SEGAN and the hyper-parameters with which it was trained and evaluated.

\subsection{Data Set}

To evaluate the effectiveness of the SEGAN, we resort to the data set by Valentini et~al.~\cite{valentiniinvestigating}. We choose it because it is open and available\footnote{\url{http://dx.doi.org/10.7488/ds/1356}}, and because the amount and type of data fits our purposes for this work: generalizing on many types of noise for many different speakers. The data set is a selection of 30~speakers from the Voice Bank corpus~\cite{veaux2013voice}: 28~are included in the train set and 2~in the test set. 

%A subset of 28~speakers (14~male and 14~female) are from the same English accent region, and the other 56 (28~male and 28~female) have different accent regions from Scotland and United States. %There are approximately 400~sentences available per speaker. 
%A subset of 28~speakers from the same English accent region is taken for %training and 2 for testing. 

To make the noisy training set, a total of 40~different conditions are considered~\cite{valentiniinvestigating}: 10~types of noise (2~artificial and 8~from the Demand database~\cite{thiemann2013diverse}) with 4~signal-to-noise ratio (SNR) each (15, 10, 5, and 0\,dB). %There are 2 artificially-generated noise types and 8 recorded that were extracted from the first channel of 48\,kHz versions of the noise recordings in the Demand database~\cite{thiemann2013diverse}. 
%The available noise types are: artificially generated (speech-shaped and babble), domestic noise (kitchen), office noise (meeting room), public space (cafeteria, restaurant, and subway station), transportation (car and metro), and street noise (busy traffic intersection).
%\begin{itemize}
%\item Artificially generated: (1) speech-shaped and (2) babble
%\item Domestic noise: inside a kitchen
%\item Office noise: meeting room
%\item Public space: (1) cafeteria, (2) restaurant and (3) subway station
%\item Transportation: (1) car and (2) metro
%\item Street noise: busy traffic intersection
%\end{itemize}
There are around 10~different sentences in each condition per training speaker. 
To make the test set, a total of 20~different conditions are considered~\cite{valentiniinvestigating}: 5~types of noise (all from the Demand database) with 4~SNR each (17.5, 12.5, 7.5, and 2.5\,dB). %The chosen test noises are: domestic noise (living room), office noise (office space), transportation (bus), and street noise (open area cafeteria and public square). 
%\begin{itemize}
%\item Domestic noise: living room
%\item Office noise: office space
%\item Transportation: bus
%\item Street noise: (1) open area cafeteria and (2) public square
%\end{itemize}
%Therefore there are 20~different noisy conditions in the test set (5~noise types with 4~SNRs each), which means there are around 20~different sentences per speaker in each condition. %For more details on how the signals were mixed please check the original paper. 
There are around 20~different sentences in each condition per test speaker. Importantly, the test set is totally unseen by (and different from) the training set, using different speakers and conditions.

\subsection{SEGAN Setup}

\begin{figure}[t]
\centering
\includegraphics[width=1\linewidth]{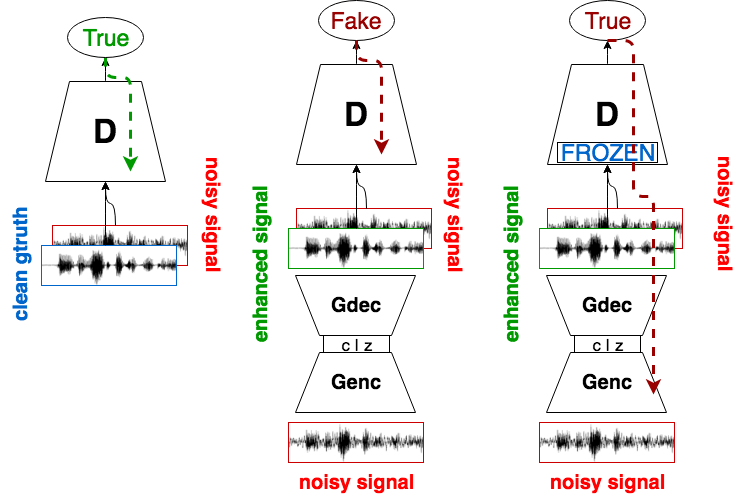}
\caption{\label{fig:GAE_process}Adversarial training for speech enhancement. Dashed lines represent gradient backprop.}
\end{figure}

The model is trained for 86~epochs with RMSprop~\cite{Tieleman12COURSERA} and a learning rate of 0.0002, using an effective batch size of 400. %The batching was parallelized in 4~GPUs to speed up the training. The batch size was picked to be 100 (an effective size of 400 with an averaging of the resulting gradients from the individual batches sent to the GPUs). 
%We used NVIDIA Titan X GPUs and a multiple-threaded loader for the data samples to avoid slowing down the training during data reading. The total time per epoch was around 20\,min. 
We structure the training examples in two pairs (Fig.~\ref{fig:GAE_process}): the real pair, composed of a noisy signal and a clean signal ($\tilde{\ve{x}}$ and $\ve{x}$), and the fake pair, composed of a noisy signal and an enhanced signal ($\tilde{\ve{x}}$ and $\hat{\ve{x}}$). %Following the training procedure described in Sec.~\ref{sec:gan}, D learns what are the appropriate characteristics to tell the difference between the clean signal and the corrupted one. On the other hand, when the fake pair is shown, we tell D to differentiate it as an invalid enhancement of the signal. This way, when G is updated to fool D, G should be correcting those mistakes that clearly show its fake behavior, with the final objective of generating more clean-alike outputs during the iterative process. 
To adequate the data set files to our waveform generation purposes, we down-sample the original utterances from 48\,kHz to 16\,kHz. During train, we extract chunks of waveforms with a sliding window of approximately one second of speech (16384~samples) every 500\,ms (50\%~overlap). During test, we basically slide the window with no overlap through the whole duration of our test utterance and concatenate the results at the end of the stream. In both train and test, we apply a high-frequecy preemphasis filter of coefficient 0.95 to all input samples (during test, output is correspondingly deemphasized).

Regarding the $\lambda$ weight of our $L_1$ regularization, after some experimentation, we set it to 100 for the whole training. We initially set it to 1, but we observed that the G loss was two orders of magnitude under the adversarial one, so the $L_1$ had no practical effect on the learning. Once we set it to 100, we saw a minimization behavior in the $L_1$ and an equilibrium behavior in the adversarial one. As the $L_1$ got lower, the quality of the output samples increased, which we hypothesize helped G being more effective in terms of realistic generation.% that made both networks stronger as the adversarial process continued.

Regarding the architecture, G is composed of 22~one-dimensional strided convolutional layers of filter width 31 and strides of $N=2$. The amount of filters per layer increases so that the depth gets larger as the width (duration of signal in time) gets narrower. The resulting dimensions per layer, being it samples $\times$ feature maps, is 16384$\times$1, 8192$\times$16, 4096$\times$32, 2048$\times$32, 1024$\times$64, 512$\times$64, 256$\times$128, 128$\times$128, 64$\times$256, 32$\times$256, 16$\times$512, and 8$\times$1024. There, we sample the noise samples $\ve{z}$ from our prior 8$\times$1024-dimensional normal distribution $\mathscr{N}(\ve{0},\ve{I})$. As mentioned, the decoder stage of G is a mirroring of the encoder with the same filter widths and the same amount of filters per layer. However, skip connections and the addition of the latent vector make the number of feature maps in every layer to be doubled. %The dimensions coming out of every layer are 8$\times$2048, 16$\times$1024, 32$\times$512, 64$\times$512, 8$\times$256, 256$\times$256, 512$\times$128, 1024$\times$128, 2048$\times$64, 4096$\times$64, 8192$\times$32, and 16384$\times$1.

The network D follows the same one-dimensional convolutional structure as G's encoder stage, and it fits to the conventional topology of a convolutional classification network. The differences are that (1)~it gets two input channels of 16384~samples, (2)~it uses virtual batch-norm~\cite{salimans2016improved} before LeakyReLU non-linearities with $\alpha=0.3$, and (3)~in the last activation layer there is a one-dimensional convolution layer with one filter of width one that does not downsample the hidden activations (1$\times$1 convolution). The latter (3)~reduces the amount of parameters required for the final classification neuron, which is fully connected to all hidden activations with a linear behavior. %(no sigmoid is placed there, as we follow the LSGAN approach). 
This means that we reduce the amount of required parameters in that fully-connected component from $8\times1024=8192$ to 8, and the way in which the 1024 channels are merged is learnable in the parameters of the convolution.

All the project is developed with TensorFlow~\cite{abadi2016tensorflow}, and the code is available at \url{https://github.com/santi-pdp/segan}. We refer to this resource for further details of our implementation. A sample of the enhanced speech audios is provided at \url{http://veu.talp.cat/segan}. 

\section{Results}
\label{sec:results}

%\subsection{Qualitative Examples}

%The results show that the proposed method can enhance speech in a wide conditions of noise. Figure~\ref{fig:preference} shows the waveform and spectrogram of the sentence \emph{We were surprised to see}. The top plot shows the clean signal and the middle plot shows the signal with additive noise from a cafeteria. The bottom plot shows the waveform generated by the generative network. I can be observed how the noise of low frequencies (till $\approx$ 4 kHz are significantly attenuated, even if it contains formant tracks of background voices. This is more effective in segments were the signal does not have low frequency content (silence and fricatives). The high frequency noise is attenuated but still present and should be investigated.\toni{Any idea? Related with nn. topology?} Some of the samples generated are provided at the web \url{veu.talp.cat/segan}.

%\begin{figure}[t]
%\centering
%  \caption{From top to bottom, waveform and spectrogram of clean, noisy (cocktail noise), and enhanced speech corresponding to the sentence ``We were surprised to see''.} 
%  \includegraphics[width=1\linewidth]{f1c.png}\\[5pt]
%  \includegraphics[width=1\linewidth]{f1n.png}\\[5pt]
%  \includegraphics[width=1\linewidth]{f1s.png}
%  \label{fig:preference}
%\end{figure}

\subsection{Objective Evaluation}

To evaluate the quality of the enhanced speech, we compute the following objective measures (the higher the better). %All measures are defined as predictors of the mean opinion score (MOS) that would be obtained in subjective tests, and all go from 1 (bad quality/high distortion/intrusive noise) to 5 (excellent quality/no degradation/not noticeable noise). 
All metrics compare the enhanced signal with the clean reference of the 824~test set files. They have been computed using the implementation included in~\cite{Loizou2013Book}, and available at the publisher website\footnote{\url{https://www.crcpress.com/downloads/K14513/K14513_CD_Files.zip}}.
\begin{itemize}[noitemsep,topsep=0pt,parsep=0pt,partopsep=0pt,leftmargin=15pt]
\item PESQ: Perceptual evaluation of speech quality, using the wide-band version recommended in ITU-T P.862.2~\cite{ITUT:P862.2} (from --0.5 to 4.5).
\item CSIG: Mean opinion score (MOS) prediction of the signal distortion attending only to the speech signal~\cite{Hu2008} (from 1 to 5).
\item CBAK: MOS prediction of the intrusiveness of background noise~\cite{Hu2008} (from 1 to 5).
\item COVL: MOS prediction of the overall effect~\cite{Hu2008} (from 1 to 5).
\item SSNR: Segmental SNR~\cite[p. 41]{quackenbush88} (from 0 to $\infty$).
\end{itemize}

%PESQ was proposed to estimate the MOS of speech processed through networks. 
%It is also correlated with the overall quality of enhanced signal~\cite{Hu2008}. 
%However, the recommendation ITU-T P.835~\cite{ITUT:P835} for evaluating systems that include noise suppression algorithms proposes separate rating scales to estimate the subjective quality of the speech signal alone (SIG), the background noise alone (BAK) and overall quality (OVL). We have included three composite measures proposed  by Hu and Loizou~\cite{Hu2008} (CSIG, CBAK, and COVL) that consist on linear regression of other objective measures (PESQ, cepstral distance, Itakura-Saito distance, etc.) to maximize correlation with subjective measures. All metrics compare the enhanced signal with the clean reference. They have been computed using the implementation included in~\cite{Loizou2013Book} and available at the publisher website\footnote{\url{https://www.crcpress.com/downloads/K14513/K14513_CD_Files.zip}}.

\begin{table}[t]
  \caption{Objective evaluation results comparing the noisy signal and the Wiener- and SEGAN-enhanced signals. %All the metrics try to predict the MOS, from 1 (bad) to 5 (good).
 }
  \label{tab:objective}
  \centering
  \begin{tabular}{lcccc}
    \toprule
    \bf Metric & \bf Noisy & \bf Wiener & \bf SEGAN\\
    \midrule
    PESQ&   1.97&  2.22&  2.16\\
    CSIG&   3.35&  3.23&  3.48\\
    CBAK&   2.44&  2.68&  2.94\\
    COVL&   2.63&  2.67&  2.80\\
    SSNR&   1.68&  5.07&  7.73\\
  \bottomrule
  \end{tabular}
\end{table}

Table~\ref{tab:objective} shows the results of these metrics. To have a comparative reference, it also shows the results of these metrics when applied directly to the noisy signals and to signals filtered using the Wiener method based on a priori SNR estimation~\cite{wiener_as96}, as provided in~\cite{Loizou2013Book}. It can be observed how SEGAN gets slightly worse PESQ. %However, PESQ, originally designed to assess the quality of telephone speech, may not be the best-suited metric to evaluate speech enhancement tasks. 
However, in all the other metrics, which better correlate with speech/noise distortion, SEGAN outperforms the Wiener method. It produces less speech distortion (CSIG) and removes noise more effectively (CBAK and SSNR). Therefore, it achieves a better tradeoff between the two factors (COVL). 

\subsection{Subjective Evaluation}

A perceptual test has also been carried out to compare SEGAN with the noisy signal and the Wiener baseline. For that, 20~sentences were selected from the test set. As the database does not indicate the amount and type of noise for each file, the selection was done by listening to some of the provided noisy files, trying to balance different noise types. Most of the files have low SNR, but a few with high SNR were also included.

A total of 16~listeners were presented with the 20~sentences in a randomized order. For each sentence, the following three versions were presented, also in random order: noisy signal, Wiener-enhanced signal, and SEGAN-enhanced signal. For each signal, the listener rated the overall quality, using a scale from 1 to 5. In the description of the 5~categories, they were instructed to pay attention to both the signal distortion and the noise intrusiveness (e.g.,~\emph{5=excellent: very natural speech with no degradation and not noticeable noise}). Listeners could listen to each signal as many times as they wanted, and were asked to pay attention to the comparative rate of the three signals. 

\begin{table}[t]
 \caption{Subjective evaluation results comparing the noisy signal and the Wiener- and SEGAN-enhanced signals. %MOS in scale from 1 (bad) to 5 (excellent) of 19 listeners and 20 test sentences per listener.
 }
 \label{tab:subjective}
 \centering
 \begin{tabular}{cccc}
   \toprule
   \bf Metric  &\bf Noisy  &\bf Wiener &\bf SEGAN\\\midrule
   MOS     &2.09  &2.70  &3.18 \\
   \bottomrule
 \end{tabular}
\end{table}

\begin{figure}[t]
 \caption{CMOS box plot (the median line in the SEGAN--Wiener comparison is located at 1). Positive values mean that SEGAN is preferred.}
 \label{fig:cmos}
\centering
\begin{tikzpicture}
  \begin{axis}[
  	width=0.8\linewidth, height=3.2cm,
    boxplot/draw direction=x,
    ytick={1,2},     yticklabels={{\small SEGAN--Noisy},{\small SEGAN--Wiener}},
%     boxplot/draw direction=y,
%     xtick={1,2},     xticklabels={{\footnotesize SEGAN--Noisy},{\small SEGAN--Wiener}},
]
%segan vs noise
%(-3);0;1;2;(5)
    \addplot+[boxplot prepared={
      lower whisker=-3, lower quartile=0,
      median=1, upper quartile=2,
      upper whisker=5, average=1.09,
      every median/.style={ultra thick},
   }]
    coordinates {};
%segan vs wiener
%(-1.5);0;0;1;(2.5)
    \addplot+[boxplot prepared={
      lower whisker=-1.5, lower quartile=0,
      median=1, upper quartile=1,
      upper whisker=2.5, average=0.47,
      every median/.style={ultra thick},
      }
     ]
    coordinates {};
  \end{axis}
\end{tikzpicture}
\end{figure}
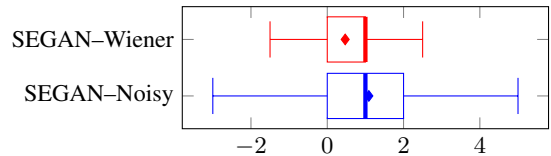

In Table~\ref{tab:subjective}, it can be observed how SEGAN is preferred over both the noisy signal and the Wiener baseline. However, as there is a large variation in the SNR of the noisy signal, the MOS range is very large, and the difference between Wiener and SEGAN is not significant. However, as the listeners compared all the systems at same time, it is possible to compute the comparative MOS (CMOS) by subtracting the MOS of the two systems being compared. Fig.~\ref{fig:cmos} depicts this relative comparison. We can see how the signals generated by SEGAN are preferred. More specifically, SEGAN is preferred over the original (noisy) signal in 67\% of the cases, while the noisy signal is preferred in 8\% of the cases (no preference in 25\% of the cases). With respect to the Wiener system, SEGAN is preferred in 53\% of cases and Wiener is preferred in 23\% of the cases (no preference in 24\% of the cases).

\section{Conclusions}
\label{sec:conclusions}

In this work, an end-to-end speech enhancement method has been implemented within the generative adversarial framework. The model works as an encoder-decoder fully-convolutional structure, which makes it fast to operate for denoising waveform chunks. The results show that, not only the method is viable, but it can also represent an effective alternative to current approaches. %This work is a first step towards a generalized end2end speech generation and manipulation framework. 
Possible future work involves the exploration of better convolutional structures and the inclusion of perceptual weightings in the adversarial training, so that we reduce possible high frequency artifacts that might be introduced by the current model. Further experiments need to be done to compare SEGAN with other competitive approaches. 

%As a possible future line of research we can explore improved designs for the convolutional structures, specially at lowest layers, where the least level of abstraction is usually learned. Introducing some prior knowledge in the filters design might be beneficial there, specially to get rid of some high frequency artifacts that are sometimes introduced with the current structure. Another option is to add some perceptual-based filtering at the raw audio level. 

\section{Acknowledgements}

This work was supported by the project TEC2015-69266-P (MINECO/FEDER, UE).

%Work supported by the Spanish Ministerio de Econom{\'i}a y Competitividad and European Regional Developmend Fund, contract TEC2015-69266-P (MINECO/FEDER, UE).

\clearpage

\bibliographystyle{IEEEtran}
\bibliography{mybib,bibjoan}

\end{document}